\documentclass{article}

\usepackage[preprint]{neurips_2023}
\usepackage[utf8]{inputenc} % allow utf-8 input
\usepackage[T1]{fontenc}    % use 8-bit T1 fonts
\usepackage{url}            % simple URL typesetting
\usepackage{booktabs}       % professional-quality tables
\usepackage{amsfonts}       % blackboard math symbols
\usepackage{nicefrac}       % compact symbols for 1/2, etc.
\usepackage{microtype}      % microtypography
\usepackage{xcolor}         % colors
\usepackage{multirow}
\usepackage[hidelinks]{hyperref}
\usepackage{graphicx}
\usepackage{xspace}
\usepackage{geometry}
\usepackage{changepage}
\usepackage{arydshln}

\usepackage{titlesec}

\titleformat{\part}[display]   % "display" option for a new page
  {\normalfont\large\bfseries} % Font settings: large, bold, centered
  {\partname\ \thepart}        % Label format
  {1pt}                       % Space between the label and title text
  {}                           % Before-code (empty here)

\newcommand{\danube}{\mbox{\textit{H2O-Danube3}}\xspace}
\newcommand{\llm}{\mbox{\textit{H2O-Danube3-4B}}\xspace}
\newcommand{\llmtiny}{\mbox{\textit{H2O-Danube3-500M}}\xspace}
\newcommand{\llmtinychat}{\mbox{\textit{H2O-Danube3-500M-Chat}}\xspace}
\newcommand{\llmchat}{\mbox{\textit{H2O-Danube3-4B-Chat}}\xspace}

\title{H2O-Danube3 Technical Report}
\author{
Pascal Pfeiffer
$\quad$Philipp Singer
$\quad$Yauhen Babakhin \\
$\quad$\textbf{Gabor Fodor} 
$\quad$\textbf{Nischay Dhankhar}
$\quad$\textbf{Sri Satish Ambati}\\
H2O.ai \\
\texttt{\{firstname.lastname, sri\}@h2o.ai}
}
\date{}

\begin{document}

\maketitle

\section{Abstract}

% In our effort to grow the ecosystem of permissive open-source foundation models, we publish a new
% set of models called  \emph{H2O-Danube3}.  %\newline 

We present \emph{H2O-Danube3}, a series of small language models consisting of \llm, trained on $6T$ tokens and \llmtiny,  trained on $4T$ tokens.
Our models are pre-trained on high quality Web data consisting of primarily English tokens in three stages with different data mixes before final supervised tuning for chat version.
The models exhibit highly competitive metrics across a multitude of academic, chat, and fine-tuning benchmarks.
Thanks to its compact architecture, \danube can be efficiently run on a modern smartphone, enabling local inference and rapid processing capabilities even on mobile devices.
We make all models openly available under Apache~2.0 license further democratizing LLMs to a wider audience economically.

\textbf{Danube3 model collection:}\newline \url{https://huggingface.co/collections/h2oai/h2o-danube3-6687a993641452457854c609} 

\section{Introduction}

Small language models have taken a pivotal place in today's open source language model landscape particularly aiming at efficient inference on consumer hardware and edge devices also allowing for full offline applications. Additionally, smaller models have proven to be particularly useful after fine-tuning them for specific tasks, such as sequence classification, question answering, or token classification even outpacing previously used encoder/decoder models such as those stemming from BERT and its derivatives \cite{devlin2018bert, he2020deberta}. 

We extend previous research in this area \cite{liu2024mobilellm, biderman2023pythia, zhang2024tinyllama, zhang2022opt, bai2023qwen, stablelm, h2odanube18b, phi3_2024} and present \danube, a series of small language models consisting of \llm, trained on $6T$ tokens and \llmtiny, trained on $4T$ tokens based on incremental research and training efforts~\cite{h2odanube18b}.
In this report, we present an overview of the models, detailing their architecture, training procedures, and fine-tuning processes. We offer extensive evaluations using a diverse range of benchmarks, encompassing both standard academic metrics, chat benchmarks, and fine-tuning benchmarks.

Results show that \danube exhibits competitive benchmarks across all dimensions, expanding the repertoire of open source small language models. We hope our work can further democratize language models to a wider audience and that our models can play a pivotal role for various use cases such as (1) chatbot applications, (2) RAG applications, (3) fine-tuning for specific use cases such as classification, (4) research or (5) on-device offline applications. To demonstrate the potential, we also present \emph{H2O AI Personal GPT}\footnote{\url{https://h2o.ai/platform/danube/personal-gpt/}}, an iOS application allowing to run \danube fully offline on a modern phone device.

\begin{figure}[t!]
    \centering
    \includegraphics[width=0.48\linewidth]{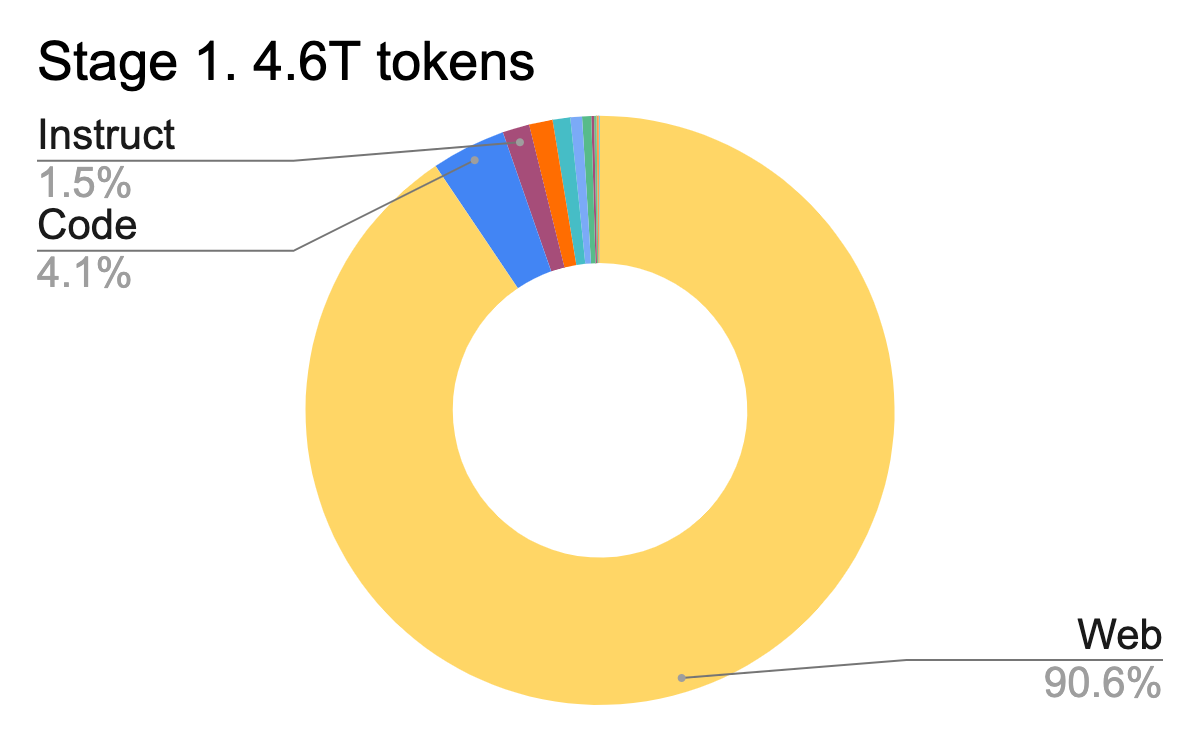}
    \includegraphics[width=0.48\linewidth]{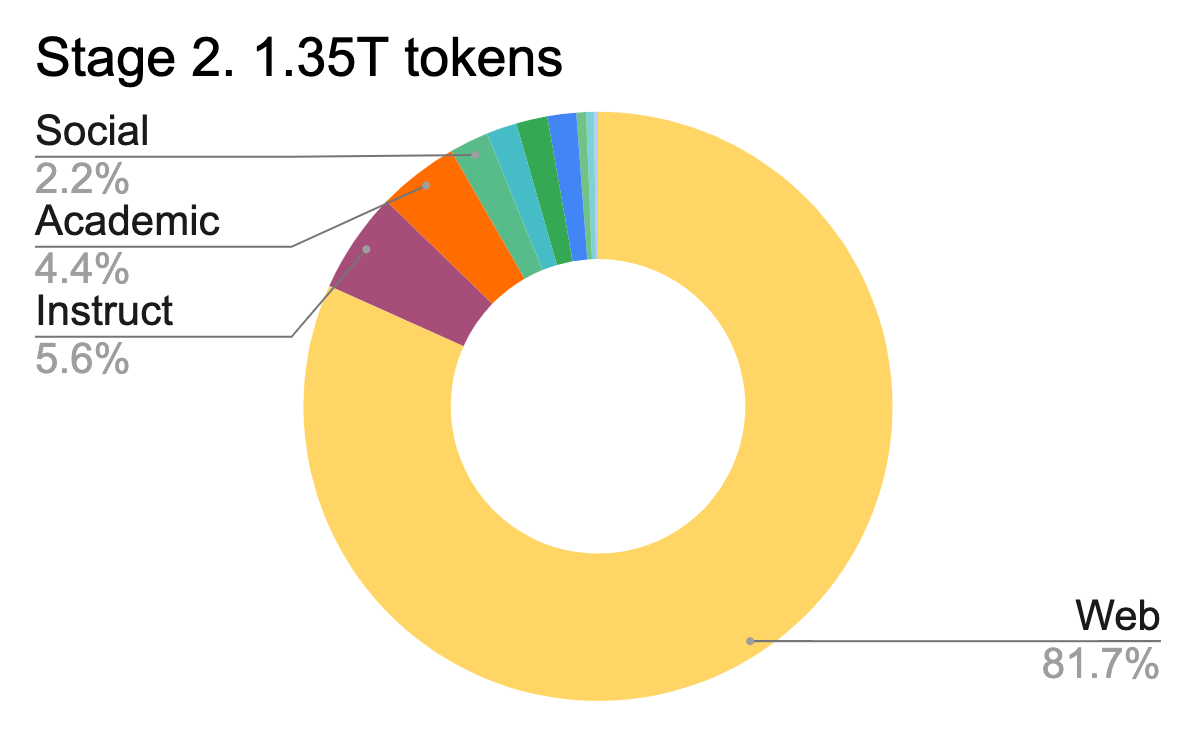}
    \includegraphics[width=0.48\linewidth]{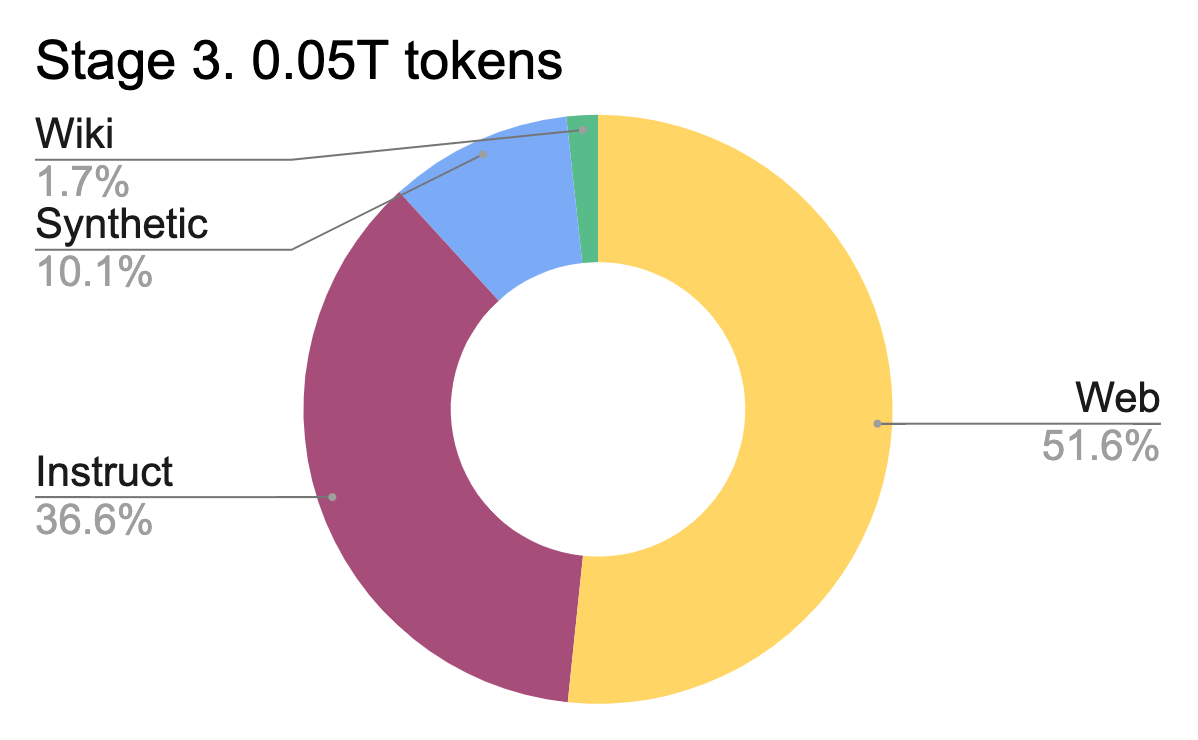}
    \caption{\textbf{Data stages for \llm.} The model is trained over three different stages with different data mixes. The first data stage consist of 90.6\% of web data which is gradually decreasing to 81.7\% at the second stage, and to 51.6\% at the third stage. The first two stages include the majority of the tokens: 4.6T and 1.35T tokens respectively, while the third stage comprises of 0.05T tokens.}
    \label{fig:data_stages}
\end{figure}

\section{Model architecture}
\label{sec:architecture}

\danube is a family of decoder only LLM models that use the general Llama model architecture adopting core principles from Llama 2 \cite{touvron2023llama} and Mistral \cite{jiang2023mistral} with custom parameters determining the shape of each layer and total parameter count. We use the Mistral tokenizer with a vocabulary size of $32,000$ and train our model up to a context length of $8,192$. We make use of Grouped Query Attention \cite{ainslie2023gqa} and optimize towards parameter and compute efficiency resulting in a wide architecture (see Table \ref{table:architecture_params}). In total, \llm consists of 3.96B trainable parameters. In addition, we release \llmtiny with 500M trainable parameters for edge devices with limited compute or for custom fine-tuning tasks that require low memory footprint or high throughput at low cost.

\begin{table}[h!]
    \centering
    \caption{Key model parameters.}
    \label{table:architecture_params}
    \begin{tabular}{lcc}
        \toprule
        Parameters & 500M & 4B \\
        \midrule
        Layers & 16 & 24 \\
        Hidden size & 1536 & 3840 \\
        Intermediate size & 4096 & 10240 \\
        Num heads & 16 & 32 \\
        Num KV heads & 8 & 8 \\
        Head size & 96 & 120 \\
        Vocab size & 32000 & 32000 \\
        RoPE theta & 100000 & 100000 \\
        \bottomrule
    \end{tabular}
\end{table}

\section{Training}
\label{sec:training}

Models are primarily trained on English text in three stages with different data mixes. At each stage, we gradually decrease the percentage of noisy web data in favor of higher quality data. The first data stage consist of 90.6\% of web data which is gradually decreasing to 81.7\% at the second stage, and to 51.6\% at the third stage. Simultaneously, the share of instruct data, Wikipedia, academic texts, synthetic texts and other higher quality textual data is increasing. The first two stages include the majority of the tokens: $4.6T$ and $1.35T$ tokens respectively ($2.8T$ and $1.15T$ tokens for \llmtiny), while third stage comprises of $0.05T$ tokens. The data distribution across stages is presented in Figure~\ref{fig:data_stages}.

We also provide chat fine-tuned versions \llmchat and \llmtinychat. We utilize \emph{H2O LLM Studio}\footnote{\url{https://github.com/h2oai/h2o-llmstudio}}, an Apache 2.0 open-source framework and no-code GUI for fine-tuning LLMs.
We tune the base model using supervised fine-tuning (SFT) on input/output conversational pairs. We mask the prompt loss, and use a custom prompt format. Hyperparameters were optimized iterating over multiple experiments.

\section{Evaluation}

In this section, we present evaluation of \danube across a variety of dimensions, focusing on (1) academic benchmarks, (2) chat benchmarks and (3) fine-tuning benchmarks.

\noindent\textbf{Academic benchmarks.} We evaluate \danube on a wide range of benchmarks and compare it with other existing open-source language models which have a similar number of parameters, specifically Qwen/Qwen1.5-4B-Chat, stabilityai/stablelm-zephyr-3b and microsoft/Phi-3-mini-4k-instruct. We also compare to our previous model h2oai/h2o-danube2-1.8b-chat. To evaluate the models, we use the Language Model Evaluation Harness framework\footnote{commit e5e5ee0cb629c9c88165292d1b4bf34623392d33} \cite{eval-harness}. \llm shows very competitive and consistent results across all reported benchmarks (see Table \ref{tab:eval_bench}). It is the best-in-class model for the knowledge based CommonsenseQA benchmark and PhysicsQA and achieves a strong accuracy of 50.14\% on the math centered benchmark GSM8K. In all other benchmarks, \llm ranks second only after Phi-3-mini-4k-instruct which is well known for its outstanding reasoning capabilities and strong benchmark scores. Notably, \llm scores over 80\% on 10-shot hellaswag benchmark, closing the gap to much larger models. The smaller \llmtiny is evaluated against the same benchmarks and compared to similar sized Qwen2-0.5B-Instruct (see Table~\ref{tab:eval_bench_500}). Our model scores highest in eight out of twelve benchmarks and we consider it a new well rounded model for this parameter count.

\begin{table}[hb]
    \centering
    \caption{\textbf{Academic benchmarks.} Academic benchmark results, compared to openly-available models of similar size and trained on general English text data. We compare the instruction fine-tuned models h2oai/h2o-danube2-1.8b-chat, h2oai/h2o-danube3-4b-chat, Qwen/Qwen1.5-4B-Chat, stabilityai/stablelm-zephyr-3b, microsoft/Phi-3-mini-4k-instruct. To evaluate the models, we use the Language Model Evaluation Harness framework \cite{eval-harness}.}
    \begin{tabular}{lc|cc|cccc}
        \toprule
        Benchmark & Metric & Danube2 & Danube3 & Qwen1.5 & StableLM & Phi3 \\
        & & 1.8B & 4B & 4B & 3B & 4B\\
        \midrule
        ARC-c & 25-shot & 43.69 & 58.96 & 42.15 & 47.70 & \textbf{63.91}\\
        Hellaswag & 10-shot & 73.91 & 80.36 & 69.46 & 73.71 & \textbf{80.62} \\
        MMLU & 5-shot & 37.83 & 54.74 & 54.03 & 44.98 & \textbf{69.43} \\
        TruthfulQA & 0-shot mc2 & 40.53 & 47.79 & 44.88 & 46.40 & \textbf{57.72}\\
        Winogrande & 5-shot & 69.30 & \textbf{76.48} & 66.22 & 65.59 & 70.80 \\
        GSM8K & 5-shot & 32.30 & 50.18 & 3.63 & 52.46 & \textbf{77.48} \\
        % \midrule
        % AgiEval (cn \& en) & 0-shot & 34.07 & 35.71 & 43.86 & 38.41 & \textbf{45.08} \\
        ARC-e & 25-shot & 74.92 & 83.84 & 73.44 & 72.10 & \textbf{87.29}\\
        BBH & 3-shot CoT & 30.39 & 38.92 & 21.03 & 36.77 & \textbf{71.42} \\
        CommonsenseQA & 3-shot & 54.30 & \textbf{79.52} & 76.09 & 75.76 & 77.81 \\
        CoQA & 0-shot F1 & 68.30 & 77.23 & 61.94 & 70.86 & \textbf{79.75} \\
        PIQA & 3-shot & 78.67 & \textbf{82.64} & 76.61 & 77.42 & 78.35 \\
        SciQ & 3-shot & 95.70 & 97.10 & 95.40 & 94.80 & \textbf{97.60} \\
        % Toxigen & 3-shot & 53.94 & 57.77 & 54.47 & 43.19 & \textbf{60.53}\\
        \midrule
        Average & & 58.32 & 68.98 & 57.07 & 63.21 & \textbf{76.01} \\
        % Average HF6 & & 49.59 & 61.41 & 46.73 & 55.14 & \textbf{69.99} \\
        \bottomrule
    \end{tabular}
    \label{tab:eval_bench}
\end{table}

\begin{table}[ht]
    \centering
    \caption{\textbf{Academic benchmarks for smaller models.} Academic benchmark results, compared to openly-available models of similar size and trained on general English text data. We compare the instruction fine-tuned models h2oai/h2o-danube3-500m-chat and Qwen/Qwen2-0.5B-Instruct. To evaluate the models, we use the Language Model Evaluation Harness framework \cite{eval-harness}.}
    \begin{tabular}{lccc}
        \toprule
        Benchmark & Metric & Danube3 & Qwen2 \\
        & & 0.5B & 0.5B \\
        \midrule
        ARC-c & 25-shot & \textbf{39.25} & 32.00 \\
        Hellaswag & 10-shot & \textbf{61.02} & 49.11 \\
        MMLU & 5-shot & 26.33 & \textbf{43.88} \\
        TruthfulQA & 0-shot mc2 & \textbf{39.96} &  39.28\\
        Winogrande & 5-shot & \textbf{61.72} & 56.99 \\
        GSM8K & 5-shot & 16.00 & \textbf{34.12} \\
        % \midrule
        % AgiEval (cn \& en) & 0-shot & 31.34 & 34.99\\
        ARC-e & 25-shot & \textbf{71.84} & 62.12 \\
        BBH & 3-shot CoT & \textbf{25.14} & 18.98 \\
        CommonsenseQA & 3-shot & 19.57 & \textbf{52.74} \\
        CoQA & 0-shot F1 & 48.02 & \textbf{54.89} \\
        PIQA & 3-shot & \textbf{74.70} & 68.72 \\
        SciQ & 3-shot & \textbf{95.40} & 92.60 \\
        % Toxigen & 3-shot & 48.83 & 50.64 \\
        % \midrule
        % Average & & 48.25 & \textbf{50.45} \\
        % Average HF6 & & 40.71 & \textbf{42.56} \\
        \bottomrule
    \end{tabular}
    \label{tab:eval_bench_500}
\end{table}

\newpage
\noindent\textbf{Chat benchmarks.}
Evaluating chat and instruct fine-tuned LLMs remains a critical challenge and can most reliably be conducted by large scale human assessment. In order to give an initial evaluation of our chat model, we resort to \emph{MT-Bench} \cite{zheng2023judging} and \emph{WildBench-v2} \cite{lin2024wildbench} benchmarks. They represent a collection of multi-turn questions across different categories followed by GPT-4 judgement which assigns a score from 1 to 10 for each model's response. Results are presented in Table~\ref{tab:mt-bench} showing that \llmchat is surpassing other similar sized models while Phi-3-mini takes the top spot. The 500M parameter version of the model \llmtinychat shows results that are comparable to Qwen2-0.5B-Instruct (see Table~\ref{tab:mt-bench-500m}).

We additionally conducted multiple internal evaluations and show them in the same tables. First, we performed a blind evaluation of chat performance (excluding the 500M models) following the idea of Chat Arena\footnote{\url{https://chat.lmsys.org/}}. This involved presenting users with random pairs of models and allowing them to prompt and vote on output preference (A better, B better, both bad, both good), followed by calculating an ELO score using MLE and bootstrapping. Second, we utilized an internal RAG (Retrieval-Augmented Generation) benchmark\footnote{\url{https://github.com/h2oai/enterprise-h2ogpte/tree/main/rag_benchmark}} to assess the models performance in question-answering tasks based on long PDF documents. We calculated an accuracy score for each model by comparing its generated responses to the ground truth answers.

\begin{table}[ht]
    \centering
    \caption{\textbf{Chat benchmarks.} \llmchat consistently performs very well across all benchmarks, surpassing other similar sized models and outperforming our previous Danube2 release, while Phi-3-mini takes the top spot.}
    \begin{tabular}{lc|cc|cccc}
        \toprule
        Benchmark & Metric & Danube2 & Danube3 & Qwen1.5 & StableLM & Phi3 \\
        & & 1.8B & 4B & 4B & 3B & 4B\\
        \midrule
        MT-Bench & Turn 1 & 6.41 & 7.28 & 6.68 & 7.10 & \textbf{8.38}\\
        MT-Bench & Turn 2 & 4.88 & 5.69 & 5.33 & 5.74 & \textbf{7.58} \\
        MT-Bench & Average & 5.64 & 6.49 & 6.00 & 6.42 & \textbf{7.98} \\
        \midrule
        WildBench-v2 & Raw score & 4.65 & 5.54 & 4.87 & 5.51 & \textbf{6.47} \\
        \midrule
        Internal Voting & ELO &  & 1531 & 1466 & 1435 & \textbf{1564} \\
        \midrule
        RAG Benchmark & Accuracy & 66.88 & \textbf{73.37} & 67.53 & 68.18 & \textbf{73.37} \\
        \bottomrule
    \end{tabular}
    \label{tab:mt-bench}
\end{table}

\begin{table}[ht]
    \centering
    \caption{\textbf{Chat benchmarks for smaller models.} The 500M parameter version of the model \llmtinychat shows results that are comparable to Qwen2-0.5B-Instruct model. In particular, they achieve a close MT-Bench average score (\llmtinychat being better in the 1st turn), while \llmtinychat produces better results on Wild-Bench benchmark.}
    \begin{tabular}{lc|cc}
        \toprule
        Benchmark & Metric & Danube3 & Qwen2 \\
         & & 0.5B & 0.5B \\
        \midrule
        MT-Bench & Turn 1 & \textbf{4.16} & 3.78 \\
        MT-Bench & Turn 2 & 2.40 & \textbf{2.76} \\
        MT-Bench & Average & \textbf{3.28} & 3.27 \\
        \midrule
        WildBench-v2 & Raw score & \textbf{3.36} & 3.11 \\
        \midrule
        Internal RAG Benchmark & Accuracy & 44.16 & \textbf{50.00} \\
        \bottomrule
    \end{tabular}
    \label{tab:mt-bench-500m}
\end{table}

\newpage
\noindent\textbf{Fine-tuning benchmarks.} A common application of small language models is their fine-tuning for various use cases to optimize performance on specific tasks. To that end, we also evaluate the different models' capability to be easily adaptable to new tasks, here focusing on text classification observed frequently across various use cases in businesses and applications.

We employ the following process for our fine-tuning benchmarks. We utilize \emph{H2O LLM Studio} offering out-of-the-box for tuning language models for classification feeding the final token logit distribution in a custom head for classification. For all models and datasets, we use the same settings with LoRA ($r=16$, $\alpha=32$) and same hyperparameters ($bs=1$, $epochs=1$, $lr=1e-4$, $diff\_lr=1e-05$, $max\_length=8192$). These settings are commonly used default settings in the field and we aim at particularly evaluating the default performance of models after tuning. 
We look at the following datasets, that all can be found on Hugging Face:

\begin{itemize}
\setlength{\itemsep}{1pt}
  \setlength{\parskip}{0pt}
  \setlength{\parsep}{0pt}
    \item stanfordnlp/imdb: binary sentence classification of imdb movie reviews
    \item knowledgator/Scientific-text-classification: classification of scientific texts into 10 most frequent classes, random 50-50 split
    \item ccdv/arxiv-classification: long context classification of arxiv papers into  11 classes
    \item ccdv/patent-classification: long context classification of patents into 9 classes
\end{itemize}

Table~\ref{tab:ft_bench} highlights the results for individual datasets and models, we always report the accuracy taking the highest probability class. We can see, that all small language models show excellent performance on text classification tasks after fine-tuning. Even small 500M parameter models can be highly competitive, exemplifying the utility of fine-tuning such models for specific use cases. Overall, \llm takes a leading spot in all benchmarks. These results can be seen as baseline results based on default hyperparameter settings. More extensive parameter sweeps would potentially improve results of all models at hand, and might also alter the order of performance. We plan on investigating such fine-tuning performance more extensively in the future. 

\begin{table}[hb]
    \centering
    \caption{\textbf{Fine-tuning benchmarks.} All models show excellent performance on various classification tasks after fine-tuning with \llm taking a top spot in most benchmarks.}
    \begin{tabular}{l|ccc|cccc}
        \toprule
        Dataset & Danube2 & Danube3 & Danube3 & Qwen1.5 & Qwen2 & StableLM & Phi3 \\
        & 1.8B & 0.5B & 4B & 4B & 0.5B & 3B & 4B\\
        \midrule
    %     arxiv & 0.825 & 0.794 & \textbf{0.830} & \textbf{0.830} & 0.820 & 0.812 & 0.796\\
    % imdb & 0.952 & 0.941 & \textbf{0.962} & 0.951 & 0.936 & 0.943 & 0.939 \\
    % patent & 0.610 & 0.582 & \textbf{0.615} & 0.611 & 0.614 & 0.598 & 0.352 \\
    % scientific & 0.819 & 0.790 & 0.846 & \textbf{0.851} & 0.817 & 0.828 & 0.807 \\

    Arxiv & 0.864 & 0.863 & 0.873 & \textbf{0.877} & 0.874 & 0.865 & 0.869\\
Imdb & 0.968 & 0.959 & \textbf{0.971} & 0.970 & 0.959 & 0.969 & 0.967 \\
Patent & 0.721 & 0.708 & \textbf{0.727} & 0.717 & 0.707 & 0.719 & 0.712 \\
Scientific & 0.868 & 0.846 & 0.872 & \textbf{0.875} & 0.855 & 0.867 & 0.870 \\
\midrule
Average & 0.855 & 0.844 & \textbf{0.861} & 0.86 & 0.849 & 0.855 & 0.854\\
        \bottomrule
    \end{tabular}
    \label{tab:ft_bench}
\end{table}

\begin{table}[ht]
    \centering
    \caption{\textbf{Model quantization.} This table summarizes different quantized versions of \llmchat showing the trade-off between size and quality of the models. Results indicate that quantized models reduced to 4 bits exhibit minimal loss in benchmark performance.}
    \begin{tabular}{lcccc}
        \toprule
        Quant method & Model size & MT-Bench & Perplexity \\
        \midrule
        F16          &  7.92 GB   &     6.43     &    6.17         \\
 Q8\_0                             &  4.21 GB   &     6.49     &    6.17      \\
Q6\_K                              &  3.25 GB   &     6.37     &    6.20          \\
 Q5\_K\_M       &  2.81 GB   &     6.25     &    6.24         \\
 Q4\_K\_M       &  2.39 GB   &     6.31     &    6.37        \\
Q3\_K\_M       &  1.94 GB   &     5.87     &    6.99      \\
 Q2\_K &  1.51 GB   &    3.71     &   9.42    \\
 
        \bottomrule
    \end{tabular}
    \label{tab:quant}
\end{table}

\newpage
\section{Model quantization}

To facilitate the use of our models on edge devices, we introduce quantized versions of \llmchat and \llmtinychat. They are available in the \danube Hugging Face collection and contain GGUF format model files that were quantized using the llama.cpp\footnote{\url{https://github.com/ggerganov/llama.cpp}} framework.

Table~\ref{tab:quant} summarizes different quantized versions of \llmchat. It shows the trade-off between size and quality of different quantization methods. Columns in the table represent quantization method, size of the model in gigabytes, MT-Bench~\cite{zheng2023judging} benchmark score, and perplexity metric on WikiText-2 dataset (as reported in a perplexity test from llama.cpp). Results suggest that we can reduce the model size by a factor of 3.3 (4-bit quantization) keeping the quality of the model almost the same, but going to 3-bit quantization already decreases the performance significantly.

%\newpage
\section{Conclusions}

We introduce \danube, a series of small language models consisting of \llm and \llmtiny released open source under Apache 2.0. Our models show competitive performance compared to popular models of similar size across a wide variety of benchmarks including (1) academic benchmarks, (2) chat benchmarks, as well as (3) fine-tuning benchmarks. \danube is built on our continuous efforts to contribute to the growing ecosystem of open source small language models. We are confident that our models can play a pivotal role in a wide range of applications, from typical chatting and fine-tuning for specific use cases to on-device offline applications on mobile phones or edge devices.

%\newpage
\small
\bibliographystyle{plain}
\bibliography{references}   % Specifies the .bib file (without the extension)

\end{document}